\documentclass[conference,10pt,a4paper,twocolumns,final]{IEEEtran}

\newlength{\flexwidth}
\usepackage{amsmath,amssymb,amsthm,fixmath}
\usepackage{xcolor}
\usepackage{siunitx}
\usepackage{cuted}
\usepackage{multirow,booktabs}
\usepackage{subcaption}
\usepackage[inline]{enumitem}
\usepackage{optidef}
\usepackage{graphicx}
\usepackage{lipsum}

\usepackage[hidelinks]{hyperref}

\usepackage{epstopdf}
\usepackage[shortcuts,acronym,automake]{glossaries}
\makeglossaries
\newacronym{5G}{5G}{Fifth Generation}
\newacronym{6G}{6G}{Sixth Generation}
\newacronym{AI}{AI}{artificial intelligence}
\newacronym{CPS}{CPS}{cyber-physical system}
\newacronym[plural={DTs}, \glsshortpluralkey={DTs}]{DT}{DT}{digital twin}
\newacronym{EI}{EI}{emergent intelligence}
\newacronym{FL}{FL}{Federated Learning}
\newacronym{IA}{IA}{industrial agent}
\newacronym{IoT}{IoT}{Internet of Things}
\newacronym{I4p0}{I4.0}{Industry 4.0}
\newacronym{IIoT}{IIoT}{industrial Internet of Things}
\newacronym[plural={MAS}, \glsshortpluralkey={MAS}]{MAS}{MAS}{multi-agent system}
\newacronym{ML}{ML}{Machine Learning}
\newacronym{PMF}{PMF}{probability mass function}
\newacronym{PSO}{PSO}{particle swarm optimization}
\newacronym{SI}{SI}{swarm intelligence}
\newacronym{UAV}{UAV}{unmanned aerial vehicle}

\newcommand{\superscript}[1]{^{\mathrm{#1}}}
\newcommand{\subscript}[1]{_{\mathrm{#1}}}

\usepackage{tcolorbox}
\tcbuselibrary{many}

\newtcbtheorem[]{alg}{Algorithm}{fonttitle=\bfseries}{alg}
\newcommand\blfootnote[1]{%
	\begingroup
	\renewcommand\thefootnote{}\footnote{#1}%
	\addtocounter{footnote}{-1}%
	\endgroup
}

\usepackage[norelsize,linesnumbered,ruled]{algorithm2e}
\makeatletter
\newcommand{\removelatexerror} {\let\@latex@error\@gobble}
\makeatother

\begin{document}

\title{Trust-Awareness to Secure Swarm\\Intelligence from Data Injection Attack}

\author{
	\IEEEauthorblockN{
		Bin~Han\IEEEauthorrefmark{1},
		Dennis~Krummacker\IEEEauthorrefmark{2},
		Qiuheng~Zhou\IEEEauthorrefmark{2},
		and~Hans~D.~Schotten\IEEEauthorrefmark{1}\IEEEauthorrefmark{2}
	}
	\IEEEauthorblockA{
		\IEEEauthorrefmark{1}RPTU Kaiserslautern-Landau, Kaiserslautern, Germany\\
		\IEEEauthorrefmark{2}German Research Center of Artificial Intelligence (DFKI), Kaiserslautern, Germany
	}
}

\bstctlcite{IEEEexample:BSTcontrol}

\maketitle

\begin{abstract}
Enabled by the emerging \ac{IA} technology, \ac{SI} is envisaged to play an important role in future \ac{IIoT} that is shaped by \ac{6G} mobile communications and \ac{DT}. However, its fragility against data injection attack may halt it from practical deployment. In this paper we propose an efficient trust approach to address this security concern for \ac{SI}.
\end{abstract}


\IEEEpeerreviewmaketitle

\glsresetall
\section{Introduction}\label{sec:introduction}
Over the past years, the technological trends of \ac{CPS} and \ac{IIoT} have raised a tide of \ac{I4p0}~\cite{SSH+2018industrial}, which has swept the world with its revolutionary use cases such as smart manufacturing, asset tracking, predictive maintenance, among others. After one decade, a variety of emerging technologies have arisen into the view and shown great potentials to push the current \ac{I4p0} one step further. The most significant ones among them are including the \ac{6G} mobile network~\cite{JHH+2021road}, \ac{ML}~\cite{ULG+2020machine}, \ac{AI}~\cite{PJL+2020industrial}, and \ac{DT}~\cite{HRS+2022impact}. As key technical enablers, they are leading us into the hallway towards the next era of industry, where numerous intelligent devices and services shall be pervasively deployed and interconnected, with humans also seamlessly included and organically integrated~\cite{Maier20216g}. 

\blfootnote{This work is supported partly by the European Commission through the H2020 project Hexa-X (GA no. 101015956), and partly by the German Federal Ministry of Education and Research (BMBF) through the project Open6GHub (GA no. 16KISK003K, 16KISK004). B. Han (bin.han@rptu.de) is the corresponding author.}

Such a vision is advocating to deploy \glspl{MAS} for a new paradigm of future smart industry: the \glspl{IA}, which is believed to be powerful against an emerging set of industrial challenges~\cite{KLR+2020industrial}. It is worth remarking that \ac{IA} provides a solid support to one promising approach of collaborative distributed intelligence: the \ac{EI}, especially the \ac{SI}~\cite{Hexa-X_D1p3}. Compared to classical centralized \ac{ML} solutions, \ac{SI} exhibits several unique advantages such as privacy, robustness, and scalability~\cite{HRS+2022impact}, and therefore becomes a potential complimentary to the emerging technology of \ac{FL} technology~\cite{Hexa-X_D7p2}.

Nevertheless, relying on information exchange among massive agents and the agents' reactions to perceived information, \ac{SI} can be fragile against data injection attacks from insiders, i.e. from the involved agents. Unfortunately, to the best of our knowledge, there has been little research effort made on the trust and security measures to enhance the robustness of \ac{SI} against such threat, except for a few highly use-case-specific studies~\cite{FJZ+2021data}. To close this gap, in this paper we propose an efficient trust-aware approach to secure \ac{SI} from data injection attack. We choose a simple use case of the classical \ac{PSO} algorithm for our study, for the reason that it is generic enough so that our contribution can be straightforwardly extended, and adopted by more complex and specific \ac{SI} techniques.

The remainder of this paper is organized as follows: We begin with Sec.~\ref{sec:problem} to present the problem under our investigation, and the conventional \ac{PSO} algorithm to solve it. Then in Sec.~\ref{sec:threat_assessment} we set up various models of the data injection attack, and assess their threat to the studied \ac{SI} use case. Afterwards, we introduce our main contributions, i.e. the trust score regression mechanism and the trust-aware \ac{PSO} approach, in Sec.~\ref{sec:trust_regression} and Sec.~\ref{sec:trust-aware_pso}, respectively. Both proposals are validated by numerical simulations. To the end, we close this paper with our conclusion and some outlooks.

\section{Problem Setup}\label{sec:problem}
\begin{figure}[!htpb]
	\centering
		\centering
		\includegraphics[width=\linewidth]{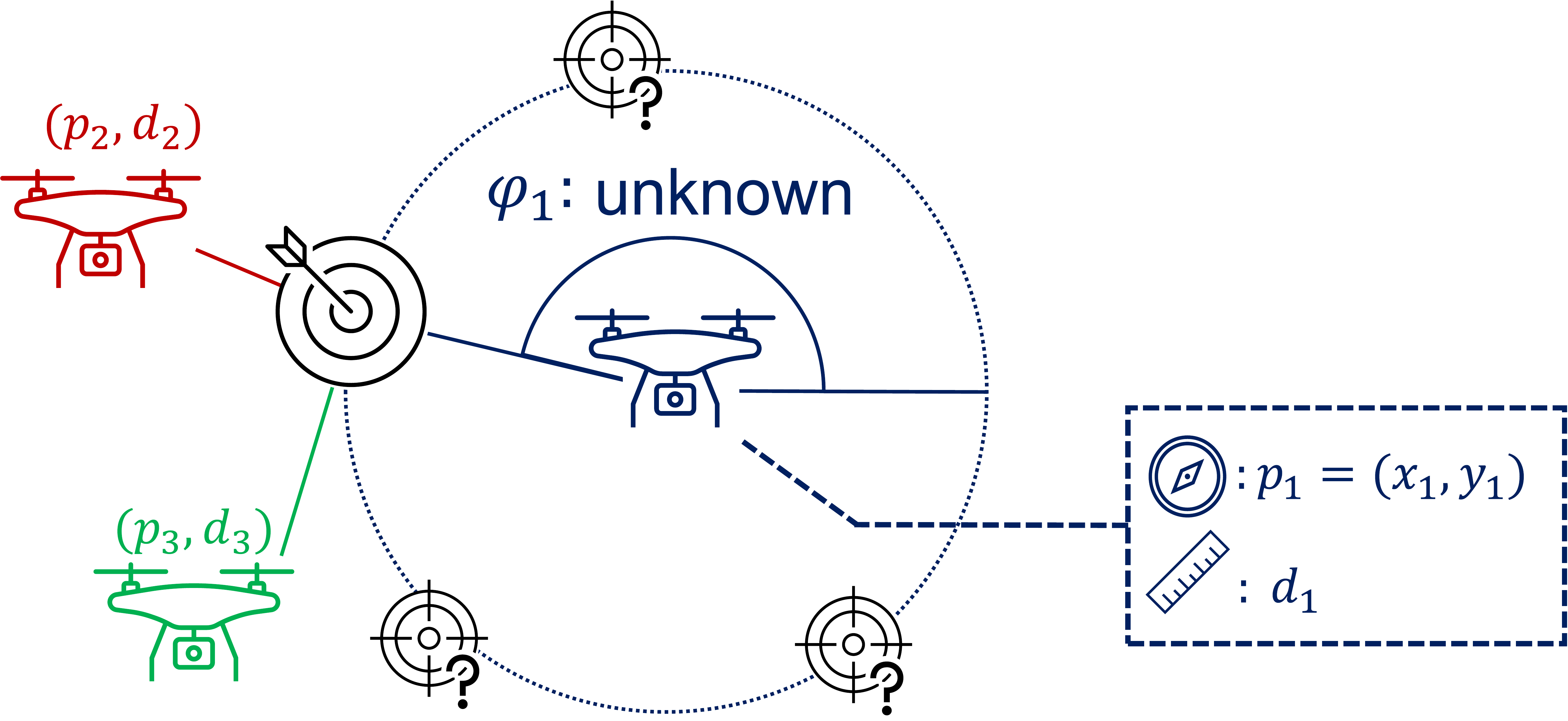}
		\caption{The investigated use case}
		\label{fig:use_case}
\end{figure}

As justified earlier, in this study we focus on one of the most typical and generic \ac{SI} methods: the \ac{PSO} algorithm. A use case of \ac{PSO}-based multi-agent joint localization, as illustrated in Fig.~\ref{fig:use_case}, is investigated. Multiple mobilized agents, denoted as a \emph{swarm} $\mathcal{I}=\{1,2,\dots I\}$, are distributed across an open area with a target at unknown position. Each agent is equipped with a positioning module, a communication device, and a non-directive sensor to measure its distance to the target. Iteratively exchanging the position and distance information with each other, agents are supposed to jointly localize the target and move towards it. To simplify the discussion we consider a two-dimensional geometric model, as an extension to the three-dimensional case will be straightforward.

In each iteration $t$, every agent $i$ obtains its accurate position $p_i^t$, but only an inaccurate distance $d_i^t$, which is the true distance $D_i^t$ modulated by a log-normal random noise:
\begin{equation}
	d_i^t=D_i^t\times 10^{\frac{n_i^t}{10}},
\end{equation}
where $n_i\sim\mathcal{N}(0,\sigma^2)$. Due to the noise, the accurate target location cannot be uniquely determined within one iteration. A conventional \ac{PSO} solution to this problem is summarized by Alg.~\ref{alg:simple_pso}. For each $(i,t)$, after measuring the data, the agent compares its current distance to target $d_i^t$ with its own record of lowest historical distance $d\superscript{best}_i$, and eventually update the latter if the former is even lower. In case of updating $d\superscript{best}_i$,  the associated position $p\superscript{best}_i$ is also updated by $p_i^t$. The swarm-best record among all agents, $\left[d\superscript{best}_\mathcal{I}, p\superscript{best}_\mathcal{I}\right]$, is also therewith checked and eventually updated. Afterwards, every $i$ corrects its previous velocity $v_i^{t-1}$ regarding its spatial offsets to both $p\superscript{best}_i$ and $p\superscript{best}_\mathcal{I}$. Two constant coefficients $c_1$ and $c_2$ are specified to adjust the long-term impact weights of $p\superscript{best}_i$ and $p\superscript{best}_\mathcal{I}$, respectively. Additionally, two random coefficients $r_1$ and $r_2$ are used to introduce short-term randomness for mitigating premature convergence. The agent speed is constrained by an upper bound $s\subscript{max}$.

\begin{algorithm}[!htpb]
	\caption{Conventional \ac{PSO} algorithm}
	\label{alg:simple_pso}
	\scriptsize
	\DontPrintSemicolon
	Input: $\mathcal{I}, s\subscript{max}, c_1, c_2, T, \left\{p_i^0: \forall i\in\mathcal{I}\right\}$\;
	Initialize: $d\superscript{best}_\mathcal{I}=+\infty,~\forall i\in\mathcal{I}: v_i^0=[0,0],~d\superscript{best}_i=+\infty$\;
	\For{$t=1:T$}{
		\For{$i\in\mathcal{I}$}{
			Update: $d_i^t$\;\label{line:update_distance}
			\If{$d_i^t<d\superscript{best}_i$}{
				$\left[d\superscript{best}_i, p\superscript{best}_i\right]\gets \left[d_i^t, p_i^t\right]$\;
				\If{$d\superscript{best}_i<d\superscript{best}_\mathcal{I}$}{
					$\left[d\superscript{best}_\mathcal{I}, p\superscript{best}_\mathcal{I}\right]\gets \left[d\superscript{best}_i, p\superscript{best}_i\right]$\label{line:swarm_best_update}
				}
			}
		}
		\For{$i\in\mathcal{I}$}{
			Generate: $[r_1,r_2]\sim \mathcal{U}^2(0,1)$\;
			$v_i^t\gets v_i^{t-1} + c_1r_1\left(p\superscript{best}_i-p_i^t\right) + c_2r_2\left(p\superscript{best}_\mathcal{I}-p_i^t\right)$\;
			\If{$\left\Vert v_i^t\right\Vert_2>s\subscript{max}$}{$v_i^t\gets{s\subscript{max}v_i^t}\left/{\left\Vert v_i^t\right\Vert_2}\right.$}
		}
		$p_i^{t+1}\gets p_i^t+v_i^t$\;
	}
\end{algorithm}

While this solution has been demonstrated in \cite{YHK+2022massive} as effective, it obviously relies on the trustworthiness of information shared among agents, and can be fragile against data injection attacks. By manipulating a minority of the involved agents to maliciously report incorrect information, an attacker is capable of misdirecting other agents' decision so that the system performance is compromised. In addition, even benevolent agents without such intention may also behave similarly and cause the same effects in unawareness, for instance, when one is equipped with a distance sensor of poor quality or in malfunction. Regardless the origin or hostility, such data reported by untrustworthy agents are threatening the system in the same way. Therefore, in this paper we do not explicitly distinguish them from each other, but generally refer to them as data injection attacks. The target of our study is to design a trust approach, which allows the \ac{PSO} algorithm to efficiently and correctly converge in presence of such attacks.

\section{Data Injection Attacks: Threat Assessment}\label{sec:threat_assessment}
\subsection{Attack Models}
Instead of consistently honestly updating $d_i^t$ to the system as supposed in Line~\ref{line:update_distance} of Alg.~\ref{alg:simple_pso}, a manipulated agent $i$ may commit a data injection attack by a certain chance:
\begin{equation}
	\alpha_i^t\sim\mathcal{B}(1,r\subscript{atk}),
\end{equation}
where $\alpha_i^t\in\{\textit{True},\textit{False}\}$ is the indicator of $i$ committing attack in iteration $t$, and $r\subscript{atk}\in(0,1]$ the attack rate. In case of attacking, $i$ replaces its raw measurement with a modified value $\tilde d_i^t$ regarding its attack model. Here we define four attack models: \begin{enumerate*}[label=\emph{\roman*)}]\item random distance, \item biased distance, \item extra distance error, and \item zero distance:
\end{enumerate*}
\begin{equation}
	d_i^t\gets\tilde d_i^t=\begin{cases}
		d_{\text{rand},i}^t,&\text{random distance;}\\
		\max\{0, d_i^t+\Delta d_i^t\},&\text{biased distance;}\\
		d_i^t \left/ \left(10^{a_i^t/10}\right)\right.,&\text{extra distance error;}\\
		0,&\text{zero distance,}
	\end{cases}
\end{equation}
where $d_{\text{rand},i}\sim\mathcal{U}_{[0,\Theta^{-1}]}$, $\Delta d_i\sim\mathcal{U}_{[-10\Theta,0]}$, $a_i\sim\mathcal{N}(0,\Theta)$, and $\Theta$ is the attacking parameter. 

\subsection{Numerical Results}
To assess the threat of data injection attack on the conventional \ac{PSO} algorithm, we conducted numerical simulations. In every individual test, $I=100$ agents were independently located regarding a uniform random distribution over a $60\times 60$~\si{\meter\squared} rectangular region, while the target is fixed at the middle of the region (but unknown to any agent). The log-normal measurement error power is set to $\sigma^2=0.1$ and the maximal agent speed $s\subscript{max}=\SI{5}{\meter/\text{round}}$. In every test, a random subset of agents $\mathcal{I}\subscript{atk}\subset\mathcal{I}$ is selected as attackers to commit a data injection attack, where $\Vert \mathcal{I}\subscript{atk}\Vert_0\sim\mathcal{U}(3,10)$. For the \ac{PSO} algorithm we set $c_1=c_2=0.5$. The same attack model is shared by all attacking agents and remains consistent throughout each individual test. To evaluate the converging performance of the \ac{PSO} algorithm, we recorded after each iteration the average agent-to-target distance among all trustworthy agents that do belong to $\mathcal{I}\subscript{atk}$. We carried out this test under different specifications of attack models and attack rates, while fixing the attack parameter $\Theta=1$. We repeated it $1000$ independent runs for every specification, with each run simulating $T=50$ iterations of the \ac{PSO} algorithm.

\begin{figure*}[!htpb]
	\centering
	\begin{subfigure}[t]{.49\linewidth}
		\centering
		\includegraphics[width=\linewidth]{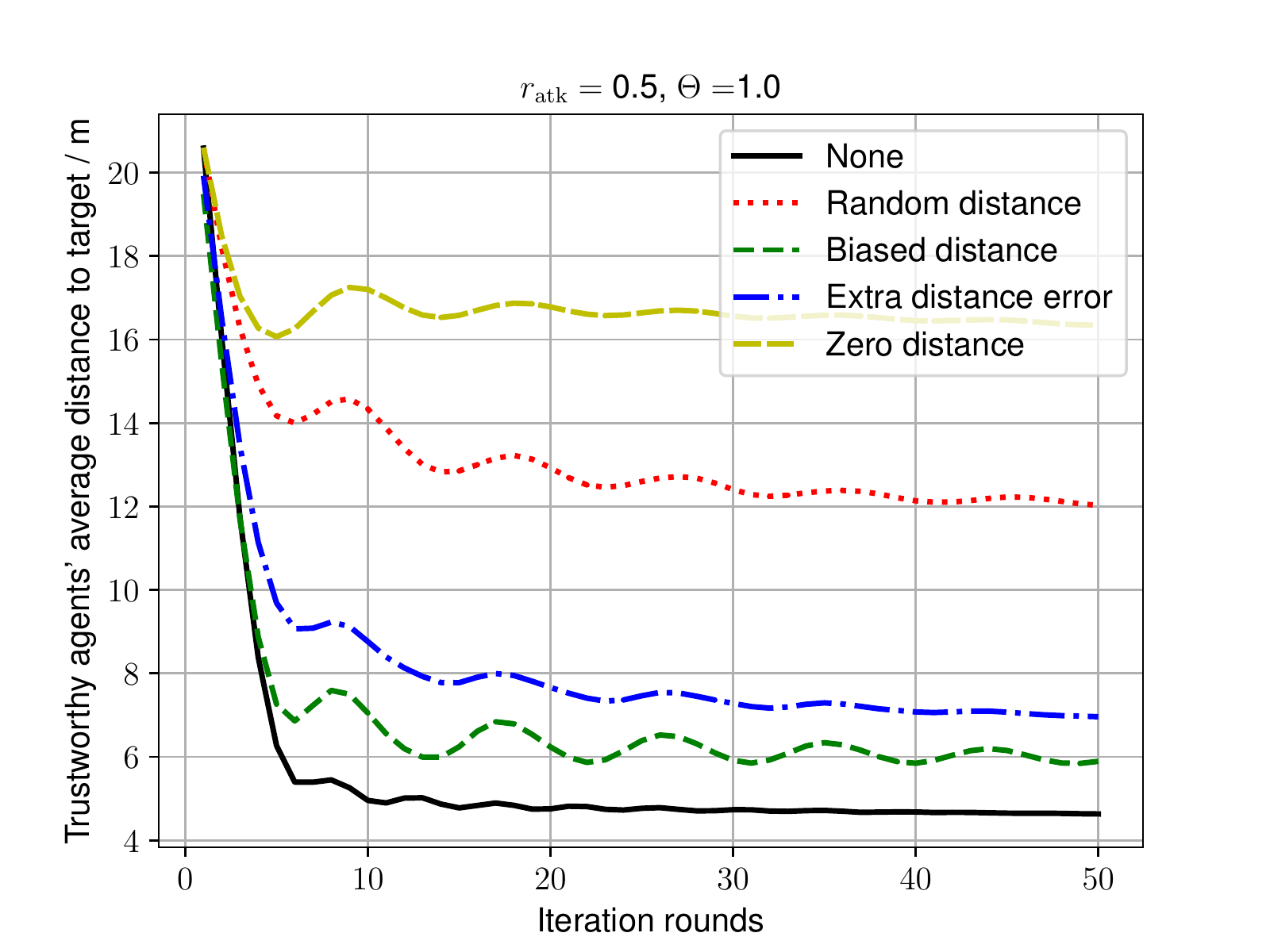}
		\caption{}
		\label{fig:threat_assmt_atk_rate_0p5}
	\end{subfigure}
	\hfill
	\begin{subfigure}[t]{.49\linewidth}
		\centering
		\includegraphics[width=\linewidth]{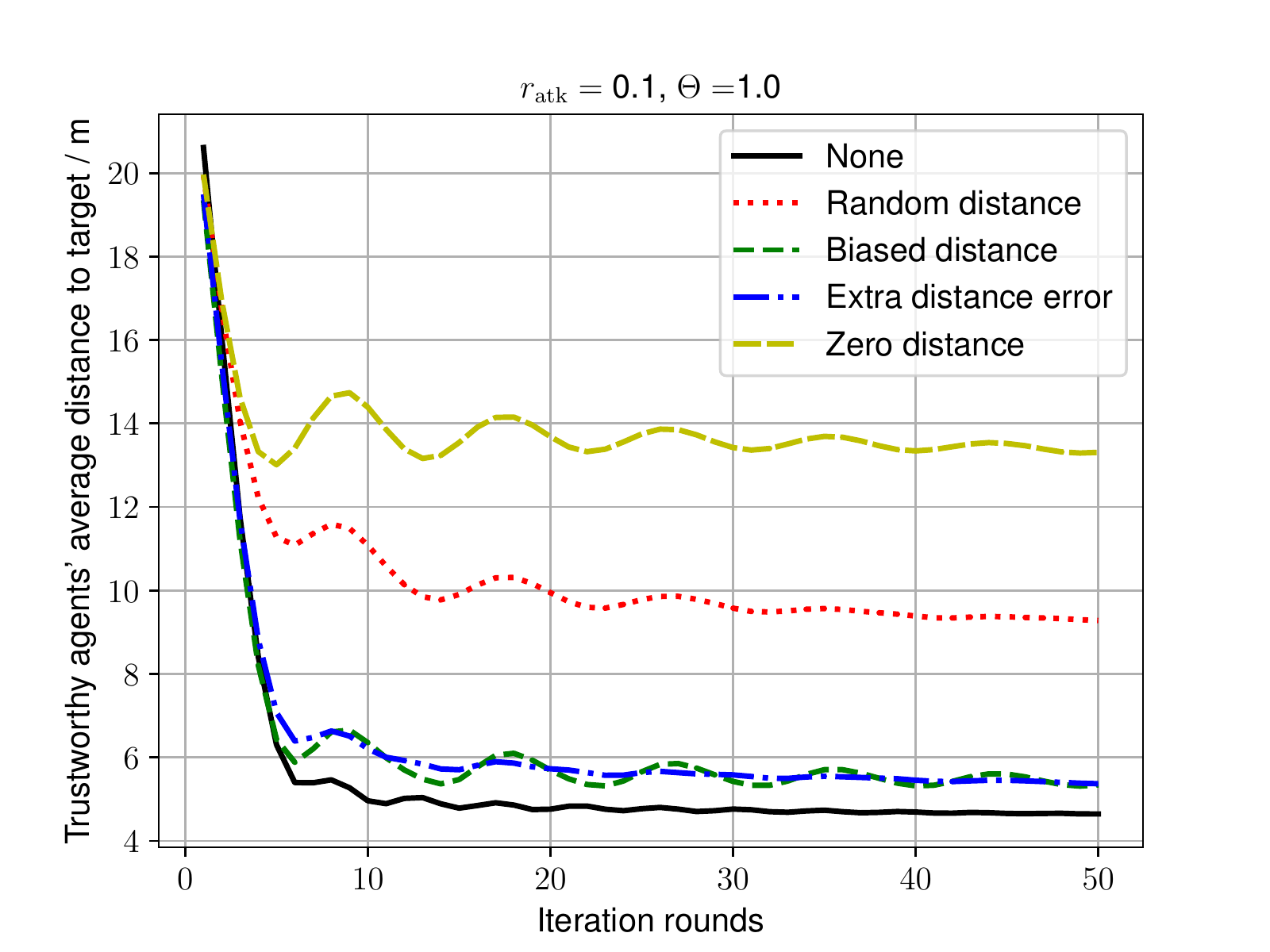}
		\caption{}
		\label{fig:threat_assmt_atk_rate_0p1}
	\end{subfigure}
	\caption{Different data injection attacks on conventional \ac{PSO}, at attack rates of
			(\subref{fig:threat_assmt_atk_rate_0p5}) 50\% and (\subref{fig:threat_assmt_atk_rate_0p1}) 10\%, respectively.}
\label{fig:threat_assessment}
\end{figure*}

The average result of the Monte-Carlo tests are comparatively illustrated in Fig.~\ref{fig:threat_assessment}. While the \ac{PSO} solution is rapidly converging to an average agent-target distance around \SI{5}{\meter} (as the curves ``None'' are showing), all kinds of data injection attacks can significantly degrade the performance at convergence, especially by reporting random or zero distance, which are capable of sufficiently failing the \ac{PSO} solution.  It deserves to be noted in particular that even under a low attack rate of $10\%$, an effective attack can be accomplished. This observation amplifies our concern to the threat of data injection attack, since the attackers can usually, if not always, effectively hide themselves from being identified by lowering the attack rate, as we will see in Sec.~\ref{sec:trust_regression}.

\section{Trust Score Regression}\label{sec:trust_regression}
To protect a multi-agent system from data injection attacks committed by insiders, there are generally two key detection challenges: \begin{enumerate*}[label=\emph{\roman*)}]
	\item the detection of data anomaly; and
	\item the detection of untrustworthy agents.
\end{enumerate*}
To address the former one, it generally relies on system-specific knowledge, which can be either model-based or empirical. The latter one, in contrary, outlines a more generic problem of evaluating the trustworthiness of an individual agent upon its historical behavior. While the two approaches are non-exclusive to each other, in this work we are focusing on the second. To omit detailed discussions about data anomaly detection while avoiding a loss of generality, we conceive a data anomaly detector with certain error rates:
\begin{align}
	p\subscript{md}&=P(\zeta_i^t~\vert~i\in\mathcal{I}\subscript{atk}~\land~\alpha_i^t),\\
	p\subscript{fa}&=P(\zeta_i^t~\vert~i\not\in\mathcal{I}\subscript{atk}~\lor~\neg\alpha_i^t),
\end{align}
where $\zeta_i^t\in\{\textit{True},\textit{False}\}$ indicates if $d_i^t$ is an anomaly, $p\subscript{md}$ and $p\subscript{fa}$ are the misdetection and false alarm rates, respectively.

\subsection{Update Models and Regression Principles}
To realize a trust mechanism with memory on the historical behavior of agents, we define a trust score for every individual agent, notified as $\rho_i^t\in[0,1]$ for agent $i$ in iteration $t$. Every agent is initialized with a certain trust $\rho_i^0$. Upon the classification of updated distance $d_i^t$ by the data anomaly detector, $\rho_i^t$ is updated in every iteration $t\in\mathbb{N^+}$. We propose three models of trust score update, namely \begin{enumerate*}[label=\emph{\roman*)}]
	\item binary,
	\item linear, and
	\item exponential,
\end{enumerate*}
respectively. More specifically, if $d_i^t$ is classified as normal, $i$ is rewarded in its trust score:
\begin{equation}
	\left.\rho_i^t\right\vert_{\neg\zeta_i^t}=\begin{cases}
		1,&\text{binary reward;}\\
		\min\{\rho_i^t + 0.05, 1\},&\text{linear reward;}\\
		\min\{2\rho_i^t, 1\},&\text{exponential reward.}\\
	\end{cases}
\end{equation}
Similarly, with $d_i^t$ classified as anomaly, $i$ is punished:
\begin{equation}
	\left.\rho_i^t\right\vert_{\zeta_i^t}=\begin{cases}
		0,&\text{binary penalty;}\\
		\max\{\rho_i^t - 0.05, 0\},&\text{linear penalty;}\\
		\frac{\rho_i^t}{2},&\text{exponential penalty.}\\
	\end{cases}
\end{equation}

Furthermore, we establish five different trust score regression strategies by flexibly combining these rewarding and punishing models: \begin{enumerate*}[label=\emph{\roman*)}]
	\item binary/binary,
	\item linear/linear,
	\item exponential/exponential,
	\item exponential/linear, and
	\item linear/exponential,
\end{enumerate*}
where the former mode in each combination stands for the reward, and the later for penalty. Especially, note that the binary/binary strategy does nothing but simply taking the raw output of data anomaly detection as the result of attacker detection, which is used only as the baseline.

The trust score regression approach can be straightforwardly combined with a threshold-based attacker detector. After each iteration, every agent is labeled upon its instantaneous trust score w.r.t. a pre-defined threshold $\rho\subscript{th}$:
\begin{equation}
	\xi_i^t=\begin{cases}
		\textit{True},&\rho_i^t<\rho\subscript{th};\\
		\textit{False},&\text{otherwise},
	\end{cases}
\end{equation}
where $\xi_i^t$ indicates if $i$ is identified as attacker at $t$.

\subsection{Numerical Results}
To evaluate the performance of our proposed trust score regression strategies, we applied each of them to the system we considered in Sec.~\ref{sec:threat_assessment}. We executed the test with zero distance attack\footnote{Indeed, since the values of $p\subscript{md}$ and $p\subscript{fa}$ are fixed regardless of the attack model, the attack model is irrelevant here and can be arbitrarily configured.} at two different rates: $10\%$ and $50\%$. For every individual specification, we repeated $1000$ independent runs of Monte-Carlo test, each lasting $T=50$ iterations of the \ac{PSO} algorithm. We set the initial trust score to $\rho_i^0=0.5$ for all $i\in\mathcal{I}$, the threshold $\rho\subscript{th}=0.382$, and $[p\subscript{md},p\subscript{fa}]=[0.5,0.05]$.

\begin{figure*}[!htpb]
	\centering
	\begin{subfigure}[t]{.49\linewidth}
		\centering
		\includegraphics[width=\linewidth]{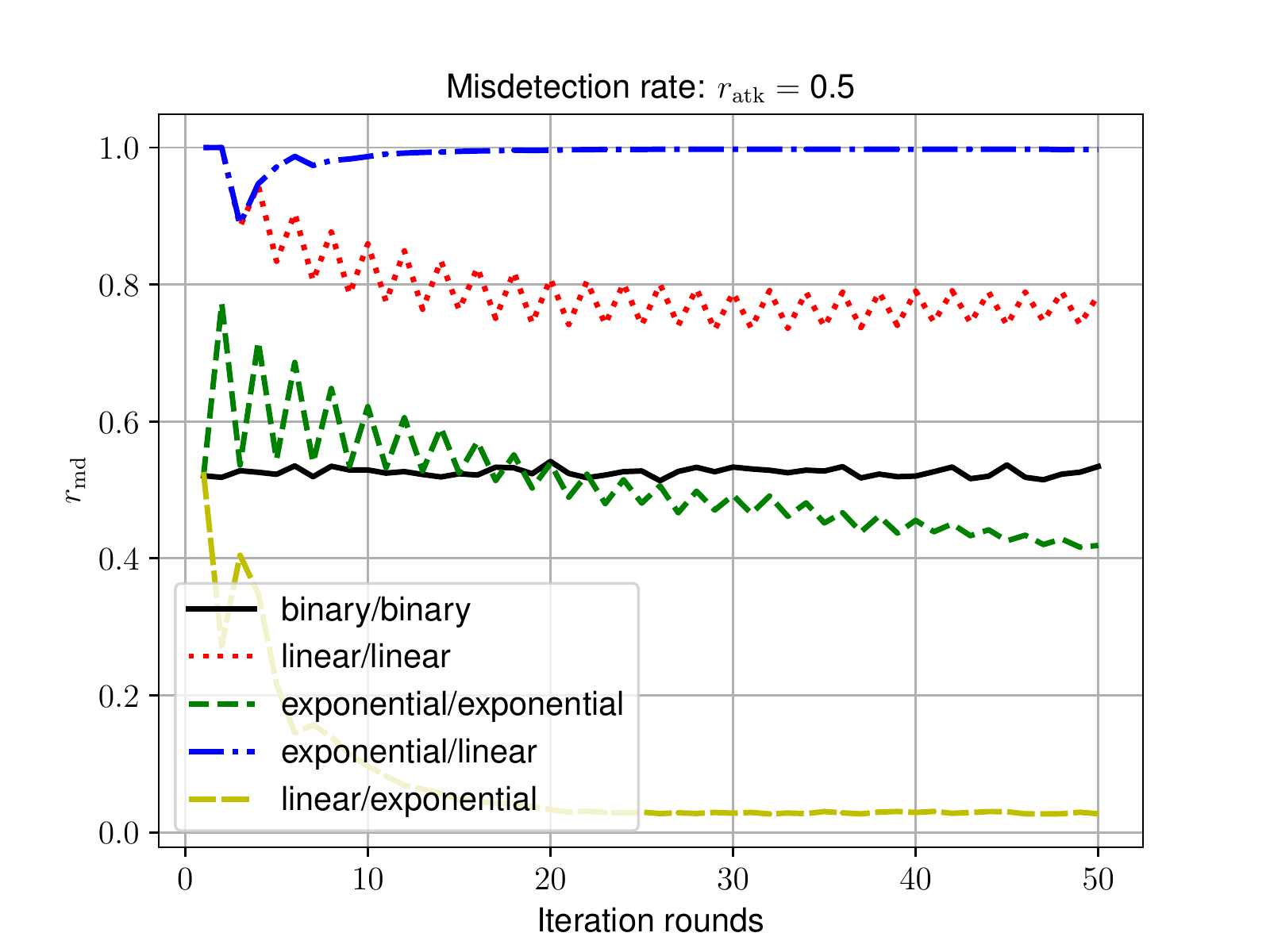}
		\caption{$r\subscript{md}$ at $50\%$ attack rate}
		\label{fig:trust_regression_mdr_atk_rate_0p5}
	\end{subfigure}
	\hfill
	\begin{subfigure}[t]{.49\linewidth}
		\centering
		\includegraphics[width=\linewidth]{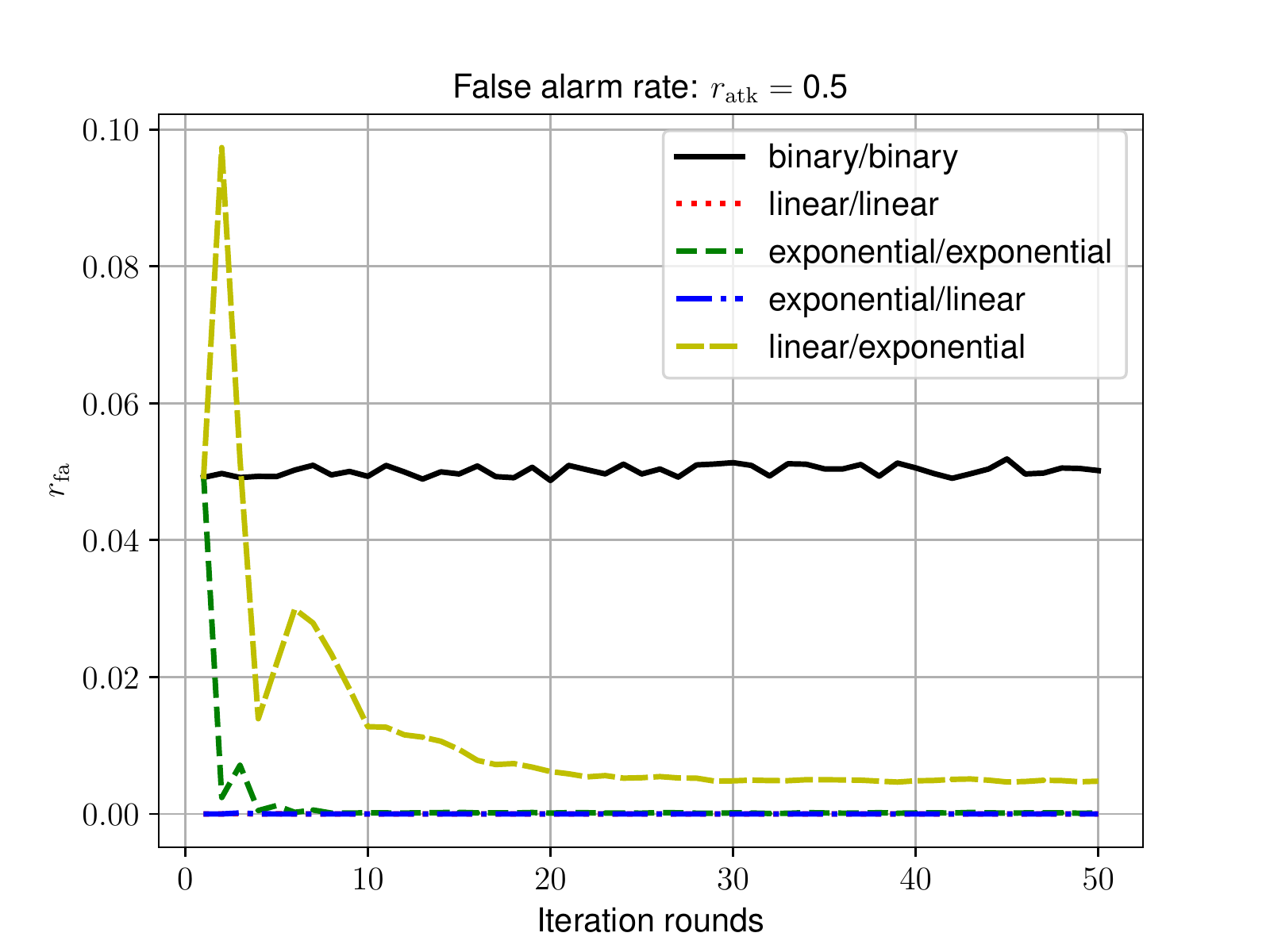}
		\caption{$r\subscript{fa}$ at $50\%$ attack rate}
		\label{fig:trust_regression_far_atk_rate_0p5}
	\end{subfigure}\\
	\begin{subfigure}[t]{.49\linewidth}
		\centering
		\includegraphics[width=\linewidth]{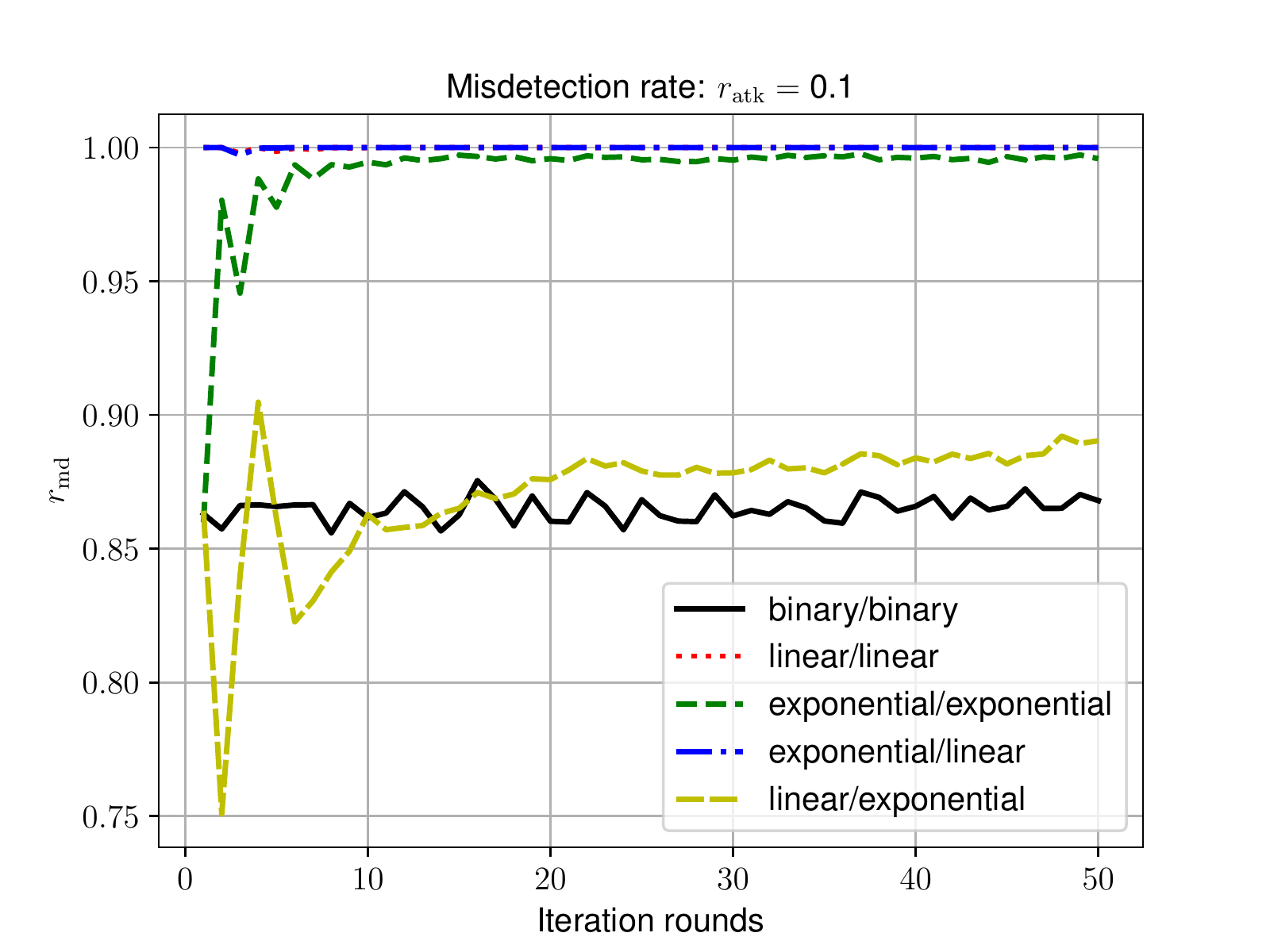}
		\caption{$r\subscript{md}$ at $10\%$ attack rate}
		\label{fig:trust_regression_mdr_atk_rate_0p1}
	\end{subfigure}
	\hfill
	\begin{subfigure}[t]{.49\linewidth}
		\centering
		\includegraphics[width=\linewidth]{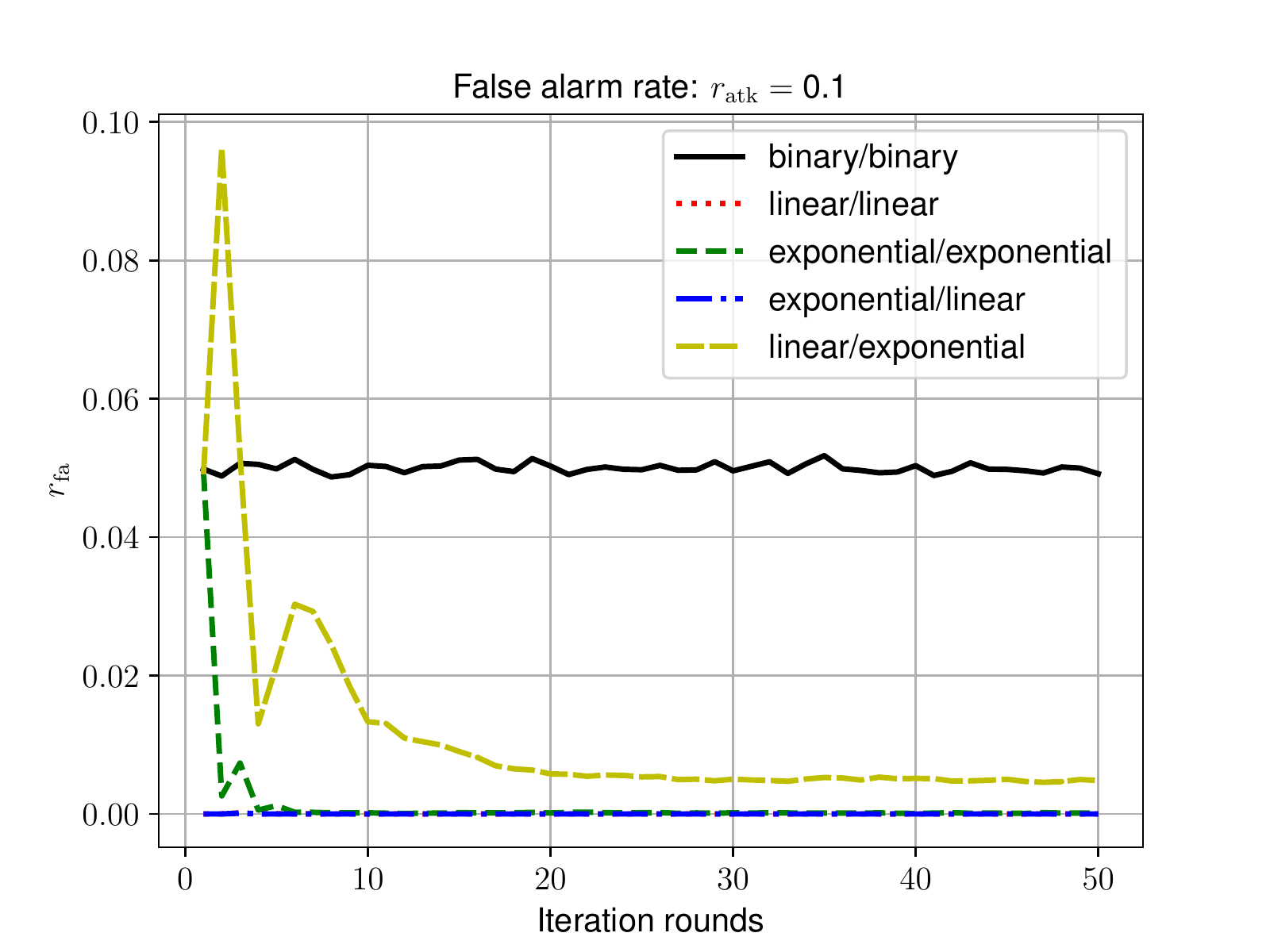}
		\caption{$r\subscript{fa}$ at $10\%$ attack rate}
		\label{fig:trust_regression_far_atk_rate_0p1}
	\end{subfigure}
	\caption{Accuracy of trust-based attacker detection with different trust score regression strategies}
	\label{fig:eva_trust_regression}
\end{figure*}

Therewith we have obtained under each specification the misdetection rate $r\subscript{md}$ and false alarm rate $r\subscript{fa}$ of the attacker detection. Remark that they shall be distinguished from those rates of the anomaly detection, i.e. $p\subscript{md}$ and $p\subscript{fa}$. As we can observe from the results shown in Fig.~\ref{fig:eva_trust_regression}, while the binary/binary baseline is exhibiting a consistent $r\subscript{fa}=p\subscript{fa}$ for an obvious reason, all four other regression strategies are capable of rapidly reducing the false alarm rate within $10$ iterations. However, even at a high attack rate of $50\%$, only the linear/exponential strategy is efficient in improving $r\subscript{md}$ w.r.t. the binary/binary baseline, while the other three are showing only limited or even negative gains. When it comes to a low attack rate of $10\%$, none of the proposed strategies can outperform the binary/binary baseline, resulting in a high $r\subscript{md}>0.85$ (or even up to $1$).

In summary, a properly designed trust score regression can help us better exploit the entire history of data anomaly detection result, and therewith improve the accuracy of attacker detection. Nevertheless, even against a detector with reasonable regression strategy, by lowing the attack rate, attackers can still manage to evade much of the detection while keeping threat to the system. Though a higher trust score threshold $\rho\subscript{th}$ will certainly help reduce $r\subscript{md}$, it essentially leads to a raised $r\subscript{fa}$ as its price. Therefore, playing around with $\rho\subscript{th}$ is no promising solution to this problem, especially since the attack rate is unknown to the system. Instead, it is rational to seek for a more advanced trust-aware mechanism.

\section{Trust-Aware Particle Swarm Optimization}\label{sec:trust-aware_pso}
\subsection{Mechanism Design}
To enhance the \ac{PSO} algorithm with a trust-awareness, the procedures of data anomaly detection, trust score regression, and trust-based attacker detection must be sequentially executed in every iteration, right after updating $d_i^t$. Therewith, the agent-reported distances can be selectively exploited regarding the attacker detection results. 

First, for all $i$ that $\xi_i^t=\textit{True}$, $d_i\superscript{best}$ shall be rejected from updating $d_\mathcal{I}\superscript{best}$. Second, since it takes time for the trust score of every agent to regress, there is a risk that an attacker $i\in\mathcal{I}\subscript{atk}$ successfully evades the trust-based attacker detection and injects a fake $\tilde{d}_i^t$ into the \ac{PSO} process as $d_\mathcal{I}\superscript{best}$ before eventually being detected. Therefore, when updating the swarm-best record $\left[d_\mathcal{I}\superscript{best}, p_\mathcal{I}\superscript{best}\right]$ with a certain entry $\left[d_i\superscript{best}, p_i\superscript{best}\right]$ (Alg.~\ref{alg:simple_pso}, Line~\ref{line:swarm_best_update}), the index $i$ of the associated agent must also be recorded as $i_\mathcal{I}\superscript{best}$ to keep track of the data source's trustworthiness, and be examined every iteration. If the source agent of the current swarm-best record is identified as attacker, the current record must be invalidated.

Moreover, regardless the specific value of $\rho\subscript{th}$, there is a fundamental and intrinsic drawback of the threshold-based attacker detector: it is incapable of dealing with ambiguous trust scores that are not significantly higher than the threshold. Data reported by an agent with such a trust score shall not be fully trusted, since there is a considerable risk that the agent is an attacker; nor shall it be simply rejected, since it still has a good chance to represent true and valuable information. Therefore, a solution beyond binary rejection is required to discriminate against data, based on the trust score of their sources. 

Thus, we propose a generic trust-aware \ac{PSO} framework as in Algorithm \ref{alg:trust_aware_pso}, where the function $\mathtt{GenBest}$ in Line~\ref{line:gen_swarm_best} updates the swarm-best record regarding agent trust scores. 

\begin{algorithm}[!htpb]
	\caption{Trust-aware \ac{PSO} algorithm}
	\label{alg:trust_aware_pso}
	\scriptsize
	\DontPrintSemicolon
	Input: $\mathcal{I},~s\subscript{max},~c_1,~c_2,~T,~\left\{p_i^0: \forall i\in\mathcal{I}\right\},~\rho\subscript{th}$\;
	Initialize: $d\superscript{best}_\mathcal{I}=+\infty,~i\superscript{best}_\mathcal{I}=1,~\forall i\in\mathcal{I}: v_i^0=[0,0],~\rho_i^0=0.5,~d\superscript{best}_i=+\infty$\;
	\For{$t=1:T$}{
		\For{$i\in\mathcal{I}$}{
			Update: $d_i^t$\;
			\If{$i\in\mathcal{I}\subscript{atk}~\land~\alpha_i^t$}{
				$d_i^t\gets \tilde{d}_i^t(\mathtt{AttackModel}, d_i^t)$
			}		
			\If{$d_i^t<d\superscript{best}_i$}{
				$\left[d\superscript{best}_i, p\superscript{best}_i\right]\gets \left[d_i^t, p_i^t\right]$\;
			}
			$\zeta_i^t\gets\mathtt{AnomalyDetection}(d_i^t)$\;
			$\rho_i^t\gets\mathtt{TrustScoreRegression}(\zeta_i^t,\rho_i^{t-1})$\;
			$\xi_i^t\gets(\rho_i^{t}<\rho\subscript{th})$
		}
		\If{$\xi_{i_{\mathcal{I}}\superscript{best}}$}{
			$d\superscript{best}_\mathcal{I}\gets+\infty$\tcp*{Reject untrustworthy swarm-best}
		}
		$\mathcal{I}\subscript{tw}^t\gets\{i\in\mathcal{I}: \xi_i^t=\textit{False}\}$\tcp*{Trustworthy agents}
		$\left[d\superscript{best}_\mathcal{I},p\superscript{best}_\mathcal{I},i_\mathcal{I}\superscript{best}\right]\gets\mathtt{GenBest}\left(\left\{d_i^t, p_i^t, \rho_i^t, i: \forall i\in\mathcal{I}\subscript{tw}^t\right\}, d_\mathcal{I}\superscript{best}, i_\mathcal{I}\superscript{best}\right)$\label{line:gen_swarm_best}\;
		\For{$i\in\mathcal{I}$}{
			Generate: $[r_1,r_2]\sim \mathcal{U}^2(0,1)$\;
			$v_i^t\gets v_i^{t-1} + c_1r_1\left(p\superscript{best}_i-p_i^t\right) + c_2r_2\left(p\superscript{best}_\mathcal{I}-p_i^t\right)$\;
			\If{$\left\Vert v_i^t\right\Vert_2>s\subscript{max}$}{$v_i^t\gets{s\subscript{max}v_i^t}\left/{\left\Vert v_i^t\right\Vert_2}\right.$}
		}
		$p_i^{t+1}\gets p_i^t+v_i^t$\;
	}
\end{algorithm}

\subsection{Swarm-Best Record Update Policies}
Regarding the specific use case under our study, we propose two policies of this update, namely \begin{enumerate*}[label=\emph{\roman*)}]
	\item hyperbolic scaling and
	\item stochastic filtering,
\end{enumerate*}
respectively.

The policy of hyperbolic scaling takes an algebraic approach: it leverages the instantaneous trust score of agents to scale their reported distances, as described in Alg.~\ref{alg:hyperbolic_scaling}. Thus, an agent with lower trust scores generally has lower chance to update the swarm-best record, even if it claims to be close to the destination.
In contrary, the policy of stochastic filtering discriminates against agents regarding their trust score in a stochastic manner, as described in Alg.~\ref{alg:stochastic_filtering}. In every iteration, a random array $\mathcal{I}\subscript{filtered}$ of agent indices is generated from the set $\mathcal{I}\subscript{tw}^t$ of agents that are currently classified as no attacker. The size of  $\mathcal{I}\subscript{filtered}^t$ equals that of $\mathcal{I}\subscript{tw}^t$, while the \ac{PMF} that an element in $\mathcal{I}\subscript{filtered}^t$ takes $i\in\mathcal{I}\subscript{tw}^t$ is proportional to $\rho_i^t$. The update to swarm-best record is thereafter executed by agents in $\mathcal{I}\subscript{filtered}^t$ instead of by those in $\mathcal{I}\subscript{tw}^t$. Thus, an agent with lower trust score has less chance to be selected into $\mathcal{I}\subscript{filtered}^t$.
Moreover, as a baseline, the simple binary rejection without additional trust-aware agent discrimination can also be represented as a specific implementation of $\mathtt{GenBest}$, as shown in Alg.~\ref{alg:binary_rejection}.

\begin{algorithm}[!htpb]
	\caption{$\mathtt{GenBest}$ with hyperbolic scaling}
	\label{alg:hyperbolic_scaling}
	\scriptsize
	\DontPrintSemicolon
	Input: $\left\{d_i^t, p_i^t, \rho_i^t, i: \forall i\in\mathcal{I}\subscript{tw}^t\right\}, d\superscript{best}_\mathcal{I}, i\superscript{best}_\mathcal{I}$\;
	\For{$i\in\mathcal{I}\subscript{tw}^t$}{
		\If{$\left({d_i^t}\left/{\rho_i^t}\right)\right.<\left({d\superscript{best}_\mathcal{I}}\left/{\rho_{i\superscript{best}_\mathcal{I}}^t}\right)\right.$}{
			$\left[d\superscript{best}_\mathcal{I},p\superscript{best}_\mathcal{I},i_\mathcal{I}\superscript{best}\right]\gets\left[d_i^t,p_i^t,i\right]$
		}
	}
	\Return $\left[d\superscript{best}_\mathcal{I},p\superscript{best}_\mathcal{I},i_\mathcal{I}\superscript{best}\right]$
\end{algorithm}

\begin{algorithm}[!htpb]
	\caption{$\mathtt{GenBest}$ with stochastic filtering}
	\label{alg:stochastic_filtering}
	\scriptsize
	\DontPrintSemicolon
	Input: $\left\{d_i^t, p_i^t, \rho_i^t, i: \forall i\in\mathcal{I}\subscript{tw}^t\right\}, d\superscript{best}_\mathcal{I}, i\superscript{best}_\mathcal{I}$\;
	Construct \ac{PMF}: $P_K(k)=\begin{cases}
		\frac{\rho_k^t}{\sum\limits_{i\in\mathcal{I}\subscript{tw}^t}\rho_i},&k\in\mathcal{I}\subscript{tw}^t;\\
		0,&\text{otherwise}
	\end{cases}$\;
	Generate: $\mathcal{I}\subscript{filtered}^t\gets\mathtt{Random}\left(\text{\ac{PMF}}=P_K,\text{size}= \left\Vert\mathcal{I}\subscript{tw}^t\right\Vert_0\right)$\;
	\For{$i\in\mathcal{I}\subscript{filtered}^t$}{
		\If{$d_i^t<d_\mathcal{I}\superscript{best}$}{
			$\left[d\superscript{best}_\mathcal{I},p\superscript{best}_\mathcal{I},i_\mathcal{I}\superscript{best}\right]\gets\left[d_i^t,p_i^t,i\right]$
		}
	}
	\Return $\left[d\superscript{best}_\mathcal{I},p\superscript{best}_\mathcal{I},i_\mathcal{I}\superscript{best}\right]$
\end{algorithm}

\begin{algorithm}[!htpb]
	\caption{$\mathtt{GenBest}$ with binary rejection}
	\label{alg:binary_rejection}
	\scriptsize
	\DontPrintSemicolon
	Input: $\left\{d_i^t, p_i^t, \rho_i^t, i: \forall i\in\mathcal{I}\subscript{tw}^t\right\}, d\superscript{best}_\mathcal{I}, i\superscript{best}_\mathcal{I}$\;
	\For{$i\in\mathcal{I}\subscript{tw}^t$}{
		\If{$d_i^t<d\superscript{best}_\mathcal{I}$}{
			$\left[d\superscript{best}_\mathcal{I},p\superscript{best}_\mathcal{I},i_\mathcal{I}\superscript{best}\right]\gets\left[d_i^t,p_i^t,i\right]$
		}
	}
	\Return $\left[d\superscript{best}_\mathcal{I},p\superscript{best}_\mathcal{I},i_\mathcal{I}\superscript{best}\right]$
\end{algorithm}

\begin{figure*}[!htpb]
	\centering
	\begin{subfigure}[t]{.49\linewidth}
		\centering
		\includegraphics[width=\linewidth]{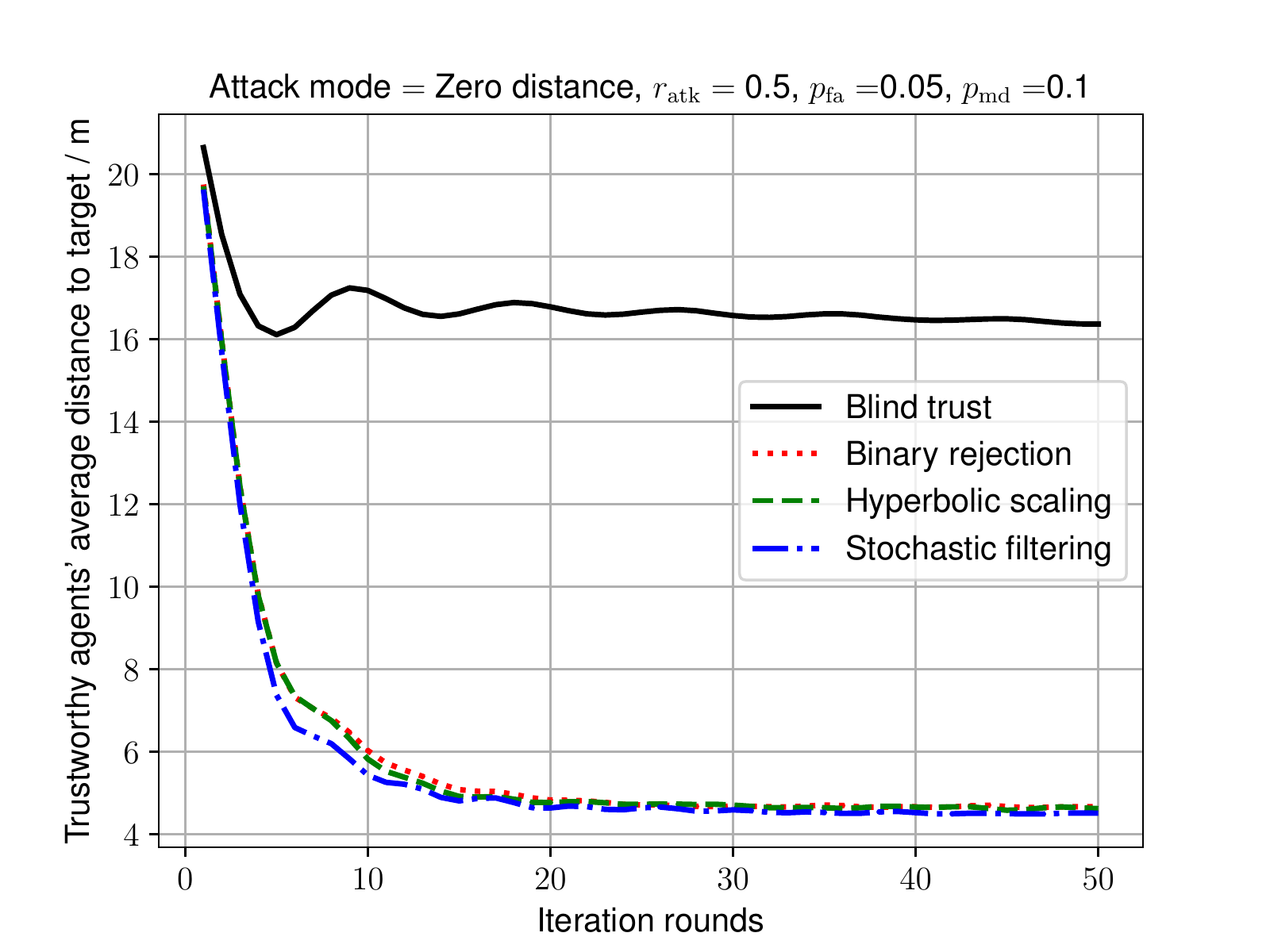}
		\caption{Zero-distance attacks by $50\%$}
		\label{subfig:benchmark_zd_atk_rate_0p5}
	\end{subfigure}
	\begin{subfigure}[t]{.49\linewidth}
		\includegraphics[width=\linewidth]{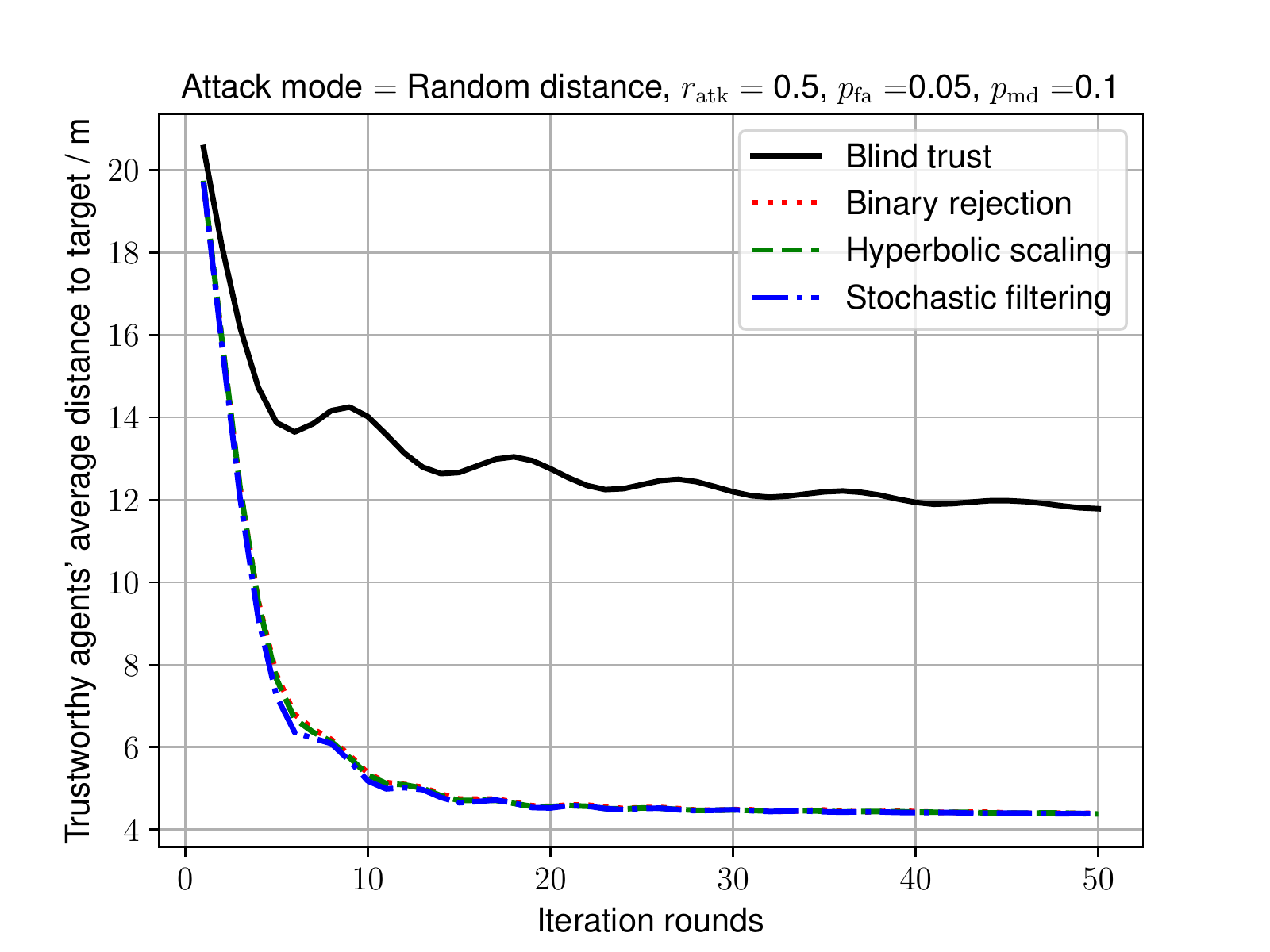}
		\caption{Random-distance attacks by $50\%$}
		\label{subfig:benchmark_rd_atk_rate_0p5}
	\end{subfigure}
	\begin{subfigure}[t]{.49\linewidth}
		\centering
		\includegraphics[width=\linewidth]{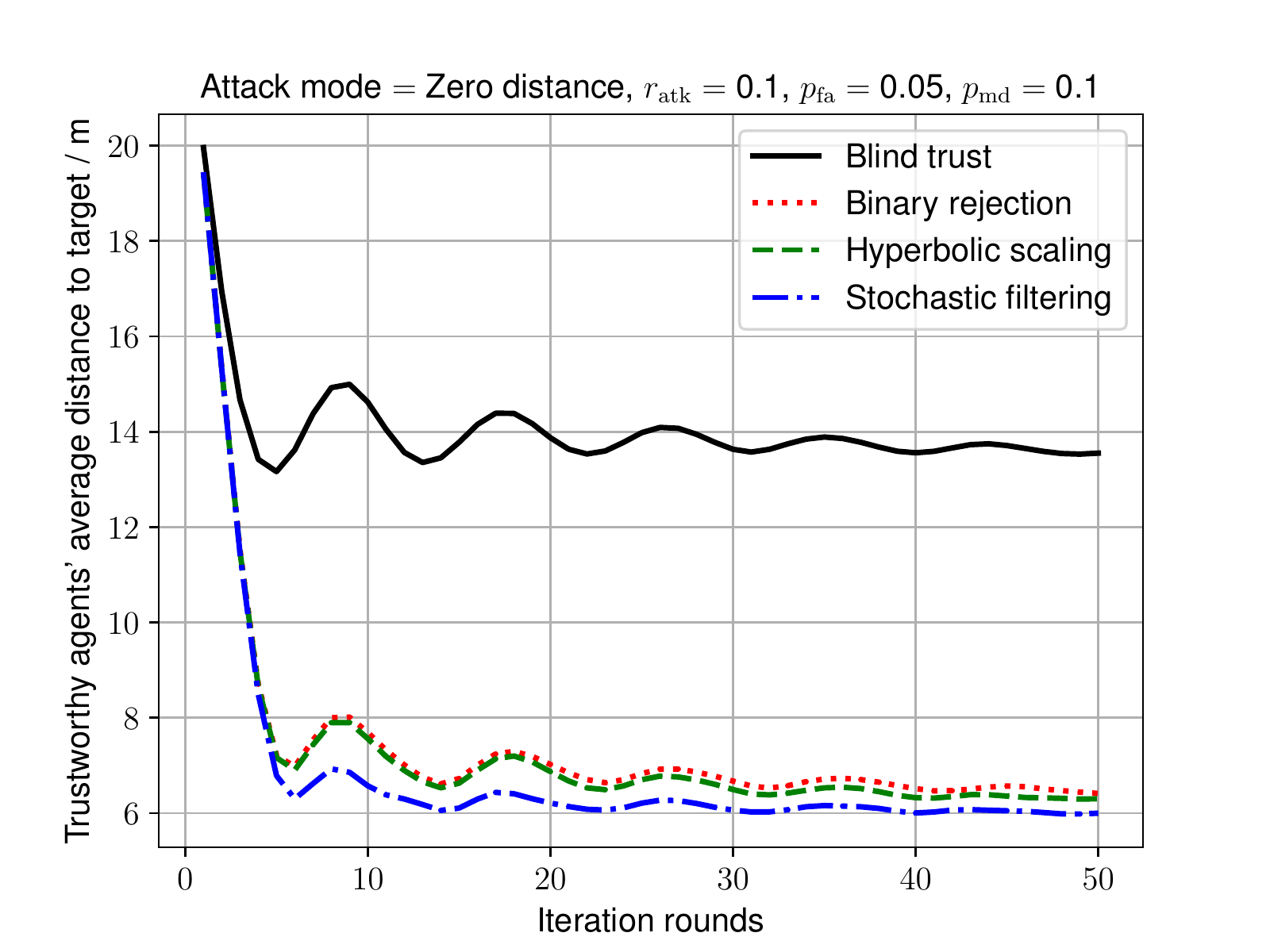}
		\caption{Zero-distance attacks by $10\%$}
		\label{subfig:benchmark_zd_atk_rate_0p1}
	\end{subfigure}
	\begin{subfigure}[t]{.49\linewidth}
		\includegraphics[width=\linewidth]{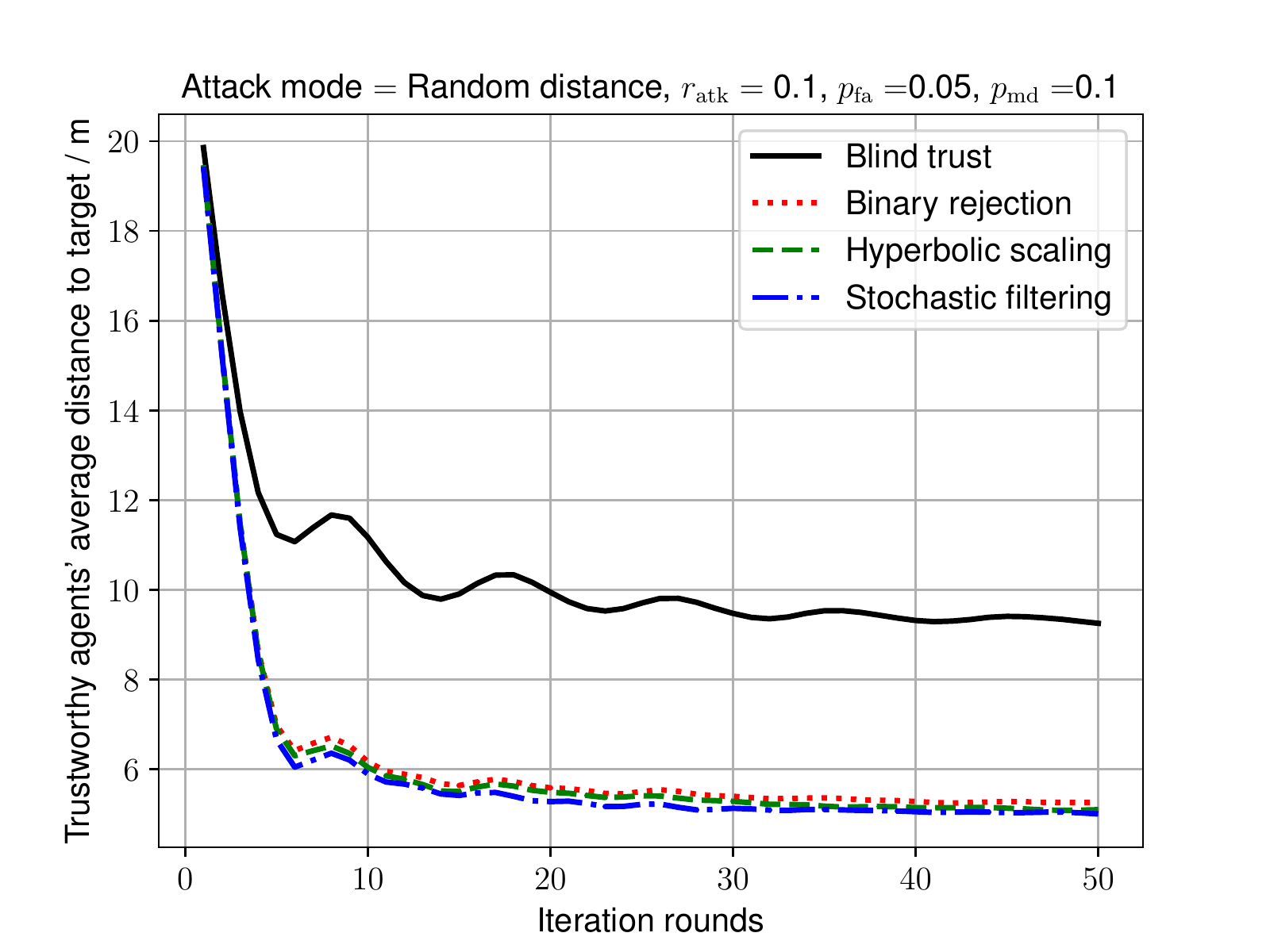}
		\caption{Random-distance attacks by $10\%$}
		\label{subfig:benchmark_rd_atk_rate_0p1}
	\end{subfigure}
	\caption{Convergence performance of trust-aware PSO with different swarm-best update policies under data injection attacks}
	\label{fig:benchmark_trust_aware_pso}
\end{figure*}

\subsection{Numerical Results}
To evaluate the performance of our proposed trust-aware \ac{PSO}, we tested it under the same system specifications as we did with the conventional \ac{PSO} in Sec.~\ref{sec:threat_assessment}. We considered the same anomaly detector as in Sec.~\ref{sec:trust_regression}, an attacker detector with the regression strategy of linear reward and exponential penalty, and initial trust score of $0.5$ for all agents. We considered the attack models of random distance and zero distance, and the attack rates of $10\%$ and $50\%$. We tested the trust-aware \ac{PSO} with both proposed policies of hyperbolic scaling and stochastic filtering, as well as the simple binary rejection. The conventional \ac{PSO} with blind trust to all agents is compared as a baseline. For every individual specification, we repeated $1000$ independent runs of Monte-Carlo test, each lasting $T=50$ iterations.

As the results in Fig.~\ref{fig:benchmark_trust_aware_pso} are showing, compared to the conventional \ac{PSO}, our proposed trust-aware \ac{PSO} algorithm significantly improves the system robustness against data-injection attacks in general. Specifically, at a high attack rate where the trust score regression is more effective in detecting attackers, all swarm-best update policies, including the simple binary rejection, are performing similarly well and sufficiently blocking fake data injections. At a low attack rate where $p\subscript{md}$ is high, the policy of stochastic filtering outperforms hyperbolic scaling, while the latter is performing only mildly better against the benchmark of binary rejection. We comprehend this performance gap between the policies as a result of their different exploitation of the trust information: by separately exploiting $\rho_i^t$ and $d_i^t$ in the processes of selective rejection and distance comparison, respectively, stochastic filtering can make better utility of the information than hyperbolic scaling, which mixes the two together and therewith lose the ability to distinguish a low distance with low trust score from a high distance with high trust score.

\section{Conclusion and Outlooks}\label{sec:conclusion}
So far, we have discussed the threat of data injection attack in \ac{SI} systems, and demonstrated our proposal to address it. According to our experiment, data injection attacks, even with a low attack rate, can significantly disrupt \ac{SI} with blind trust among agents. As a solution, our proposed trust-aware approach is proven efficient to counter such attacks.

As for the next step, it is interesting to design and implement an efficient solution of data anomaly detection for the studied problem, and evaluate our proposed approach with the specific detector. Furthermore, analytical efforts are also required in addition to enable parameter optimization of our proposed methods.





\bibliographystyle{IEEEtran}
\bibliography{references}

\end{document}